\documentclass[runningheads]{llncs}
\usepackage{multirow}
\usepackage[ruled,vlined]{algorithm2e}
\usepackage{graphicx}
\newcommand\blfootnote[1]{%
  \begingroup
  \renewcommand\thefootnote{}\footnote{#1}%
  \addtocounter{footnote}{-1}%
  \endgroup
}

\usepackage{hyperref}
\hypersetup{
    colorlinks=true,
    linkcolor=cyan,
    filecolor=magenta,      
    urlcolor=blue,
}
\begin{document}
\title{Fast Geometric Surface based Segmentation of Point Cloud from Lidar Data}
%
%
\author{Aritra Mukherjee\inst{1}\orcidID{0000-0003-3329-3287} 
Sourya Dipta Das\inst{2}\orcidID{0000-0003-1488-7418} \and
Jasorsi Ghosh\inst{2}\orcidID{0000-0002-5248-919X} \and
Ananda S. Chowdhury\inst{2}\orcidID{0000-0002-5799-3467} \and
Sanjoy Kumar Saha\inst{1}\orcidID{0000-0003-2241-5373}}

\authorrunning{A Mukherjee et al.}
%
\institute{Dept. of Computer Science and Engineering, Jadavpur University, India \\
\email{kalpurush1601@gmail.com,sks\_ju@yahoo.co.in} \and
Dept. of Electronics and Telecommunication Engineering, Jadavpur University, India\\
\email{\{dipta.math,jasorsi13,ananda.chowdhury\}@gmail.com}}
\maketitle              
\begin{abstract}
Mapping the environment has been an important task for robot navigation and  Simultaneous Localization And Mapping (SLAM). LIDAR provides a fast and accurate 3D point cloud map of the environment which helps in map building. However, processing millions of points in the point cloud becomes a computationally expensive task. In this paper, a methodology is presented to generate the segmented surfaces in real time and these can be used in modeling the 3D objects. At first an algorithm is proposed for efficient map building from single shot data of spinning Lidar. It is based on fast meshing and sub-sampling. It exploits the physical design and the working principle of the spinning Lidar sensor. The generated mesh surfaces are then segmented by estimating the normal and considering their homogeneity. The segmented surfaces can be used as proposals for predicting geometrically accurate model of objects in the robots activity environment. The proposed methodology is compared with some popular point cloud segmentation methods to highlight the efficacy in terms of accuracy and speed. \blfootnote{\textcopyright All rights reserved by Springer Nature Switzerland AG 2019}\blfootnote{B. Deka et al. (Eds.): PReMI 2019, LNCS 11941, pp. 415–423, 2019.
\newline\url{https://doi.org/10.1007/978-3-030-34869-4_45}}\blfootnote{The final publication is available at \url{https://link.springer.com/chapter/10.1007/978-3-030-34869-4_45}}

\end{abstract}
\section{Introduction}
\label{intro}
Mapping of environment in 3D is a sub-task for many robotic applications and is a primary part of Simultaneous Localization And Mapping (SLAM). For 3D structural data sensing, popular sensors are stereo vision, structured light, TOF (Time Of Flight) cameras and Lidar. Stereo vision can extract RGBD data but the accuracy is highly dependent on the presence of textural variance in the scene. Structured light and TOF cameras can extract depth information and RGBD data respectively in an accurate fashion. But these are mostly suitable for indoor uses and in a low range. Lidar is the primary choice for sensing accurate depth in outdoor scenarios over long range. Our focus in this work is on the unsupervised geometric surface segmentation in three dimensions based on Lidar data. We would like to emphasize that models built from such segmentation can be very useful for tasks like autonomous vehicle driving.

 According to Nguyen \textit{et al.}~\cite{p2}, the classic approaches in point cloud segmentation can be grouped into edge based methods \cite{p11}, region based methods \cite{p5,p19,p12,p8}, attributes based methods \cite{p10,p7,p17,p18}, model based methods \cite{p13}, and graph based methods \cite{p3,p14}. Vo \textit{et al.} \cite{p5} proposed a new octree-based region growing method with refinement of segments and Bassier \textit{et al.}. \cite{p19} improved it with Conditional Random Field. In \cite{p12,p8}, variants of region growing methods with range image generated from 3D point cloud are reported. Ioannou \textit{et al.} \cite{p7} used Difference of Normals (DoN) as a multiscale saliency feature used in a segmentation of unorganized point clouds. Himmelsbach \textit{et al.} \cite{p6} treated the point clouds in cylindrical coordinates and used the distribution of the points for line fitting to the point cloud in segments. Ground surface was recognized by thresholding the slope of those line segments. In an attempt to recognize the ground surface, Moosmann \textit{et al.} \cite{p15} built an undirected graph and characterized slope changes by mutual comparison of local plane normals. Zermas \textit{et al.} \cite{p1} presented a fast instance level LIDAR Point cloud segmentation algorithm which consisting of deterministic iterative multiple plane fitting technique for the fast extraction of the ground points, followed by a point cloud clustering methodology named Scan Line Run (SLR). On the other hand, in supervised methods, PointNet \cite{p4} takes the sliding window approach to segment large clouds with the assumption that a small window can express the contextual information of the segment that it belongs to. Landrieu \textit{et al.} \cite{p3} introduced superpoint graphs, a novel point cloud representation with rich edge features encoding the contextual relationship between object parts in 3D point clouds. This is followed by deep learning on large-scale point clouds without major sacrifice in fine details.

In most of the works, neighbourhood of a point used to estimate normal is determined by tree based search. This search is time consuming and the resulting accuracy is also limited for sparse point cloud as provided by a Lidar in robotic application. This observation acts as the motivation for us to focus on developing a fast mesh generation procedure that will provide the near accurate normal in sparse point cloud. Moreover, processing Lidar data in real-time requires considerable computational resources limiting its deployability on small outdoor robots. Hence a fast method is also in demand. 

\section{Proposed Methodology}
\label{propose}

\begin{figure}[ht]
\centering
\includegraphics[width=\textwidth]{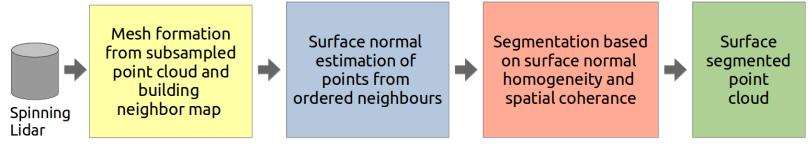}
\caption{Block diagram of the entire system, the first stage can be merged with Lidar scanning by improvising the Lidar firmware }
\label{fig:lidar_block_dia}
\end{figure}

The overall process consists of three steps as shown in Figure~\ref{fig:lidar_block_dia}. First, the point cloud is sensed by the Lidar, subsampled and the mesh is created simultaneously. Second, surface normal is calculated for node points using the mesh. Finally, surface segmentation is done by labelling the points on the basis of spatial continuity of normal with a smooth distribution.  

\begin{figure}[ht]
\centering
\includegraphics[width=\textwidth]{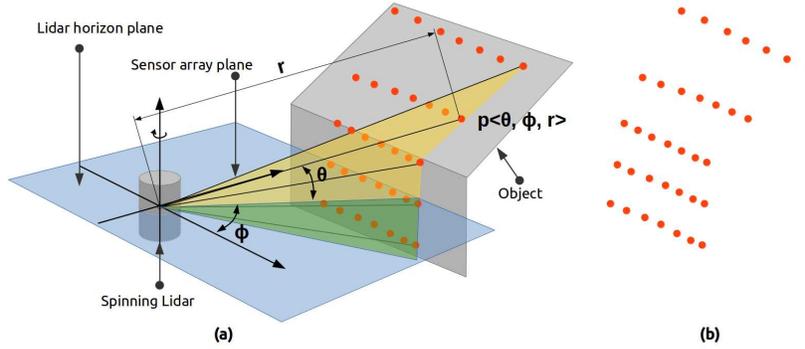}
\caption{(a) A schematic showing the formation of point cloud by Lidar and (b) the resultant point cloud }
\label{fig:lidar_op}
\end{figure}

\begin{figure}[ht]
\centering
\includegraphics[width=\textwidth]{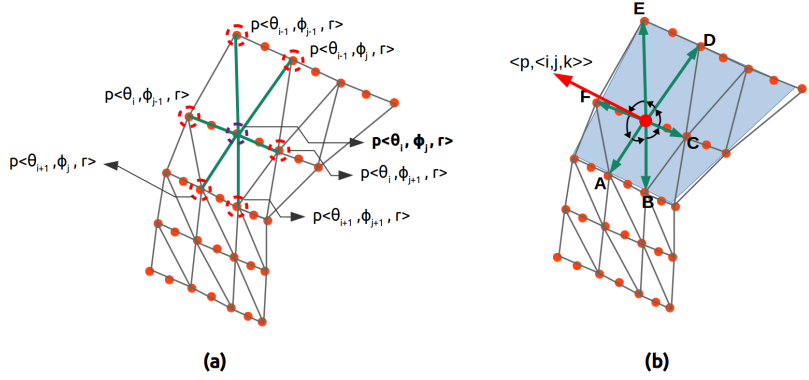}
\caption{(a) A schematic showing the formation of mesh on subsampled (factor 2) cloud with the neighbour definition of a point and (b) the normal formation from the neighbours }
\label{fig:lidar_meshnorm}
\end{figure}
The proposed methodology segments surface from point cloud obtained by spinning Lidars only. Spinning Lidars work on the principle of spinning a vertical array of divergent laser distance sensors and thus extracts point cloud in spherical coordinates. The point cloud consists of a set of coordinates $P = \{p(\theta , \phi , r)\}$ where $\theta$ is the fixed vertical angle of a sensor from the plane perpendicular to the spinning axis, $\phi$ is the variable horizontal angle due to spinning of the array and $r$ is the distance measured by the laser sensor. This form of representation is exploited by our methodology to structure the data in an ordered form the only caveat being running it for a single spin. By varying the factor of sub-sampling of $\phi$, the horizontal density of the point cloud can be varied. Figure~\ref{fig:lidar_op} shows the operational procedure of a spinning Lidar and the resultant point cloud for an object with multiple surfaces. Please note that not every point during the sweep is considered for mesh construction as noisy points too close to each other horizontally produces erroneous normal. Sub-sampling is done to rectify this error by skipping points uniformly during the spin.
\\

\noindent{\bf Mesh Construction:}
The significant novelty of this work is the fast mesh generation process that enables quick realization of subsequent steps. The mesh construction is a simultaneous process during the data acquisition stage. The connections are done during sampling in the following manner. Let a point be denoted as $p(\theta , \phi , r)$. Let the range of $\theta$ be $[\theta_0,\theta_n]$ which corresponds to $n+1$ vertical sensors in the array; and the range of $\phi$ be $[\phi_0,\phi_m]$ where $m+1$ is the number of times the sensor array is sampled uniformly during a single spin. The corresponding distance of the point is $r$ from the Lidar sensor. Let the topmost sensor in the array corresponds to angle $\theta_0$ and the count proceeds from top to bottom the vertically spinning array. Further, let first shot during the spin corresponds to $\phi_0$. The mesh is constructed by joining neighbouring points in the following manner: $p(\theta_i , \phi_j , r)$ is joined with $p(\theta_{i+1} , \phi_j , r)$, $p(\theta_{i+1} , \phi_{j+1} , r)$ and $p(\theta_i , \phi_{j+1} , r)$ for all points within range of $[\theta_0,\theta_{n-1}]$ and $[\phi_0,\phi_{m-1}]$. The points from the last vertical sensor, \textit{i.e.}, corresponding to $\theta_n$, are joined with the immediate horizontal neighbour. Thus, $p(\theta_n , \phi_j , r)$ is joined with $p(\theta_n , \phi_{j+1} , r)$ where $j$ varies from $0$ to $m-1$. For points corresponding to $\phi_m$, $p(\theta_i , \phi_m , r)$ is joined with $p(\theta_i , \phi_0 , r)$, $p(\theta_{i+1} , \phi_0 , r)$ and $p(\theta_{i+1} , \phi_m , r)$. The point $p(\theta_n , \phi_m , r)$ is joined with $p(\theta_n , \phi_0 , r)$ to create the whole cylindrical connected mesh. For all pairs the joining is done if both the points have an $r$ that is within range of the Lidar. If all the neighbours of a point is present then six of them are connected by the meshing technique instead of all eight. This is done to ensure there are no overlapping surface triangles. Figure ~\ref{fig:lidar_meshnorm}(a) shows the connectivity a point which have six valid neighbours on the mesh. The mesh is stored in a map of vectors $M=\{ <p,v> \mid p \in P, v = \{ q_n \mid q_n \in P, n\leq6,\quad and\quad q_n\quad is\quad a\quad neighbour\quad of\quad p \} \}$ where each point is mapped to the vector $v$ containing its existent neighbors in an ordered fashion. The computational complexity of the meshing stage in $O(n_{sp})$ where $n_{sp}$ is the number of sub-sampled points. As the meshing is performed on the fly with the sensor spinning the absolute time depends on the angular frequency and sub-sampling factor.
\\

\noindent{\bf Normal Estimation:}
The structured mesh created in the previous step helps toward a fast computation of normal at a point. A point forms vectors with its neighbour. Pair of vectors are taken in an ordered fashion. Normal is estimated for that point by averaging the resultant vectors formed by cross multiplication of those pairs. The ordering is performed during the mesh construction stage only. For a point $p(\theta_i , \phi_j , r)$, vectors are formed with the existing neighbours as stored in $M$ in an anti-clockwise fashion. A normal can be estimated for $p$ if its corresponding $v$ has $\vert v \vert \geq 2 $. From the neighbour vector $v$ of $p$ obtained from $M$, let $p(\theta_i, \phi_j, r)$ forms $\vec{A}$ by joining with $ p(\theta_{i+1} , \phi_j , r)$, $\vec{B}$ by joining with $ p(\theta_{i+1}, \phi_{j+1}, r)$, and so on until it forms $\vec{F}$ by joining with $ p(\theta_i, \phi_{j-1}, r)$. Then cross- multiplication of existing consecutive vectors is performed. In general, if every point exists, then $\vec{A} \times \vec{B}$, $\vec{B} \times \vec{C}$ etc. are computed ending with $\vec{F} \times \vec{A}$ to complete the circle. 
This arrangement is illustrated in Figure ~\ref{fig:lidar_meshnorm}(b). The normal is estimated by averaging all the $\hat{i},\hat{j},\hat{k}$ components of the resultant vectors individually. The normals are stored in the map $N=\{<p,n> \mid p \in P, n=\{\hat{i_p},\hat{j_p},\hat{k_p} \} \}$. Due to the inherent nature of the meshing technique, sometime points from disconnected objects get connected to the mesh. To mitigate this effect of a different surface contributing to the normal estimation,  weighted average is used. The weight of a vector formed by cross multiplication of an ordered pair is inversely proportional to the sum of lengths of the vectors in the pair. 

\begin{algorithm}[ht]
\label{algo:surfaceSeg}
\caption{Surface segmentation on normal distribution}
\SetAlgoLined
$L=\{<p,l> \mid p \in P, l=0\}$\;
$label\leftarrow 1, stack \: S\leftarrow\{\}$\;
\For{each $p \in M$}{
    \eIf{$l=0$ for $p$ in $L$}{
        $L \leftarrow <p,l\leftarrow label>$\;
        S.push(p)\;
        \While{$S \neq \{\}$}{
            $p\leftarrow$ S.pop()\;
            \For{each $q$ in $v$ corresponding to $p \in M$}{
                \If{$l=0$ for $q$ in $L$}{
                    get $\hat{i_q},\hat{j_q},\hat{k_q}$ for $q$ from $N$\;
                    \If{$\vert \hat{i_p} - \hat{i_q} \vert < I$, $\vert \hat{j_p} - \hat{j_q} \vert < J$, $\vert \hat{k_p} - \hat{k_q} \vert < K$}{
                        $L \leftarrow <q,l\leftarrow label>$\;
                        S.push(q)\;
                    }
                }
            }
        }
    }
    {search next $p$ in $M$}
}
 
\KwResult{L }
\end{algorithm}

\noindent{\bf Segmentation by Surface Homogeneity:}
Based on the normal at a point, as computed in the previous step, we now propagate the surface label. A label map $L=\{<p,l> \mid p \in P, l=0\}$ is used for this purpose. This label map stores the label of each point $p$ by assigning a label $l$. If for any point $p$, its $l=0$ denotes the point is yet to be labelled. The criteria of assigning the label of $p$ to its neighbour $q$ depends on the absolute difference of their normal components. Three thresholds $I,J,K$ are empirically set depending on the type of environment. Segment labelling is propagated by a depth first search approach as described in algorithm ~\ref{algo:surfaceSeg}. Two neighbouring points will have same label provided the absolute difference of corresponding components of their normals are within component-wise threshold. Computations of normals and mesh, as discussed earlier, generates the normal map $N$ and the mesh $M$ respectively. Subsequently algorithm ~\ref{algo:surfaceSeg} uses $N$ and $M$ to label the whole sub-sampled point cloud in an inductive fashion. Due to sub-sampling, all points in $P$ will not get a label. This issue is resolved by assigning the label of its nearest labelled point along the horizontal sweep. An optional post-processing may be arranged by eliminating segments with too few points.

\section{Experiments Results and Analysis}
\label{expr}
The proposed methodology is implemented with C++ on a linux based system with DDR4 8GB RAM, 7th generation i5 Intel Processor. Experiments were performed on a synthetic dataset which simulates Lidar point clouds. The methodology is compared with the standard region growing algorithm used in point cloud library~\cite{p9} and a region growing algorithm combined with merging for organized point cloud data~\cite{p18} and it is observed that it excels in terms of both speed and accuracy.

\begin{figure}[ht]
\centering
\includegraphics[width=\textwidth]{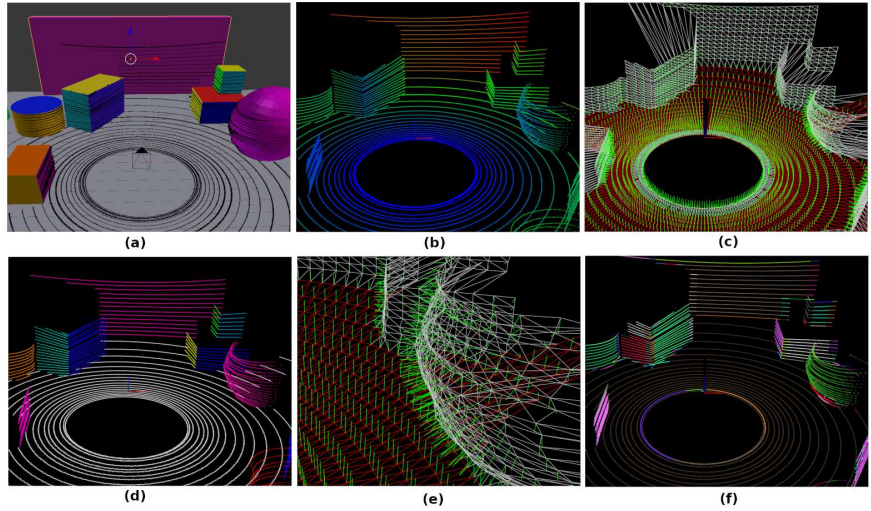}
\caption{(a) Synthetic scene with scanned point cloud overlayed (b) Point cloud with distance color coded from blue(least) to red(highest) (c) Mesh and normals with sub-sampling factor of 5 (d) Point cloud segment surface ground truth (e) A detailed look at the mesh and normals (f) Segmented point cloud by proposed methodology }
\label{fig:lidar_snaps}
\end{figure}

We have used the software ``Blensor"~\cite{p16} to simulate the output of the Velodyne 32E Lidar in a designed environment. Blensor can roughly model the scenery with the desired sensor parameters. BlenSor also provides an exact measurement ground truth annotated over the model rather than the point cloud. We have included primitive 3D model structures such as cylinder, sphere, cube, cone and combined objects placed in different physically possible orientations in the scene to simulate the real environment. Different percentage of occlusion and crowding levels is included in the model environment to test out the property of scene complexity independence, of the proposed solution. We have a total of 32 different environments with increasing order of different types of objects, occlusion, complexity of geometry and pose, and crowding levels. We have used Gaussian noise as our noise model for Lidar with zero mean and variance of $0.01$. With Velodyne 32E Lidar the total number of sensors in the spinning array is $32$ and a resolution of $0.2$ degree is set for horizontal sweep resulting in $1800$ sensor firing in one spin. Thus for our dataset the range of $\theta$ and $\phi$ are $[\theta_0,\theta_{31}]$ and $[\phi_0,\phi_{1799}]$. The output at different stages of our methodology is shown in Figure~\ref{fig:lidar_snaps}
\\

\begin{table}[ht]
\centering
\caption{Comparison of execution times (all units in milliseconds) of different competing methods.}
\label{tab:timing}
\begin{tabular}{|c|c|c|c|c|}
\hline
Method & \begin{tabular}[c]{@{}c@{}}Sampling\\  Interval\end{tabular} & \begin{tabular}[c]{@{}c@{}}Average\\  Time (in ms)\end{tabular} & \begin{tabular}[c]{@{}c@{}}Max\\  Time (in ms)\end{tabular} & \begin{tabular}[c]{@{}c@{}}Min\\  Time (in ms)\end{tabular} \\ \hline
\multirow{3}{*}{\begin{tabular}[c]{@{}c@{}}Proposed\\  Method\end{tabular}} & 5 & 54.33 & 63 & 41 \\ \cline{2-5} 
 & 10 & 35.06 & 48 & 28 \\ \cline{2-5} 
 & 15 & 25.53 & 32 & 20 \\ \hline
\multicolumn{2}{|c|}{\begin{tabular}[c]{@{}c@{}}Region \\ Growing~\cite{p9}\end{tabular}} & 275.13 & 1507 & 134 \\ \hline
\multicolumn{2}{|c|}{\begin{tabular}[c]{@{}c@{}}Region\\  Growing \\ with Merging~\cite{p18}\end{tabular}} & 368.46 & 1691 & 129 \\ \hline

\end{tabular}
\end{table}

\noindent{\bf Performance of Proposed Methodology:}
The input scene and point cloud along with the output at different stages are given in Figure~\ref{fig:lidar_snaps}. It should be mentioned that the different colors of the mesh are due to separation of ground plane on the basis of normal and is rendered for better visualization only. 
Performance is evaluated using the precision-recall and f1 score metric. As the segments may lack semantic labels, edge based comparisons were performed with overlapping of dilated edge points with ground-truths. We vary the sampling interval from $5$ to $15$ in steps of $5$, as a sampling interval of $5$ corresponds to $1$ degree of sweep of the Lidar. The thresholds for checking normal homogeneity are kept at $<\hat{i},\hat{j},\hat{k}> = <0.2,0.2,0.2>$. These values are determined empirically for scenes with standard objects on a flat surface. Our methodology is compared with ~\cite{p9} and ~\cite{p18}. The tuning parameters of the methods are kept at default settings. Table~\ref{tab:timing} shows the execution times for different methods and clearly reveals that the proposed method is much faster. In Table~\ref{tab:accuracy}, we present the accuracy values for the competing approaches. This table indicates that our solution is much more accurate. Overall, the results demonstrate that we are successful in providing a fast yet accurate solution to this complex problem.

\begin{table}[ht]
\centering
\caption{Comparison of accuracy of different competing methods}
\label{tab:accuracy}
\begin{tabular}{|c|c|c|c|c|}
\hline
Method & \begin{tabular}[c]{@{}c@{}}Sampling\\  Interval\end{tabular} & \begin{tabular}[c]{@{}c@{}}Average\\  F1 score\end{tabular} & \begin{tabular}[c]{@{}c@{}}Average \\ Precision\end{tabular} & \begin{tabular}[c]{@{}c@{}}Average\\  Recall\end{tabular} \\ \hline
\multirow{3}{*}{\begin{tabular}[c]{@{}c@{}}Proposed\\  Method\end{tabular}} & 5 & 0.7406 & 0.7616 & 0.7224 \\ \cline{2-5} 
 & 10 & 0.7147 & 0.7330 & 0.6998 \\ \cline{2-5} 
 & 15 & 0.6910 & 0.7190 & 0.6673 \\ \hline
\multicolumn{2}{|c|}{\begin{tabular}[c]{@{}c@{}}Region \\ Growing~\cite{p9}\end{tabular}} & 0.3509 & 0.4752 & 0.2804 \\ \hline
\multicolumn{2}{|c|}{\begin{tabular}[c]{@{}c@{}}Region\\  Growing \\ with Merging~\cite{p18}\end{tabular}} & 0.3614 & 0.3849 & 0.3430  \\ \hline
\end{tabular}

\end{table}

\section{Conclusion}
\label{conc}
In this work we have presented an unsupervised surface segmentation algorithm which is fast, accurate and robust to noise, occlusion and different orientations of the surface with respect to the Lidar. This work serves as the first step for mapping environments with geometric primitive modelling in SLAM applications for unmanned ground vehicles. In future, supervised classifier can be utilized for segment formation on data collected by Lidar on a real environment. Thereafter, the surface segments will enable the model generation of 3D objects.

%
%
%
 \bibliographystyle{splncs04}
 \bibliography{main}

\end{document}